%% file: conference_101719.tex
\UseRawInputEncoding
\documentclass[conference]{IEEEtran}
\IEEEoverridecommandlockouts
\usepackage{cite}
\usepackage{amsmath,amssymb,amsfonts}
\usepackage{graphicx}
\usepackage{textcomp}
\usepackage{xcolor}
\def\BibTeX{{\rm B\kern-.05em{\sc i\kern-.025em b}\kern-.08em
    T\kern-.1667em\lower.7ex\hbox{E}\kern-.125emX}}
\begin{document}

\title{\textbf{\textit{EREBUS}}:End-to-end Robust Event Based Underwater Simulation\\

}

\author{Hitesh Kyatham
$^{1}$*, Arjun Suresh$^{1}$*, Aadi Palnitkar$^{1}$*, Yiannis Aloimonos$^{1}$%
\thanks{* Equal Contributors}
\thanks{$^{1}$Maryland Robotics Center, University of Maryland, College Park, MD 20742, USA
        {\tt\small \{hkyatham,apalnitk, arjsur,jyaloimono\}@umd.edu}}%
}

\maketitle

\begin{abstract}
The underwater domain presents a vast array of challenges for roboticists and computer vision researchers alike, such as poor lighting conditions and high dynamic range scenes. In these adverse conditions, traditional vision techniques struggle to adapt and lead to suboptimal performance. 

Event-based cameras present an attractive solution to this problem, mitigating the issues of traditional cameras by tracking changes in the footage on a frame-by-frame basis. In this paper, we introduce a pipeline which can be used to generate realistic synthetic data of an event-based camera mounted to an AUV (Autonomous Underwater Vehicle) in an underwater environment for training vision models. We demonstrate the effectiveness of our pipeline using the task of rock detection with poor visibility and suspended particulate matter, but the approach can be generalized to other underwater tasks. 
\end{abstract}

\section{Introduction}
\input{sections/introduction}

\section{Related Work}

\input{sections/related_work}

\section{Methodology}
\input{sections/methods}

\section{Conclusions and Future Work}
\input{sections/future_work}

\bibliographystyle{ieeetr}
\bibliography{references}

\end{document}

%% file: sections/introduction.tex
Underwater environments present unique challenges for robotic perception, including low-light conditions, high dynamic range scenes, and turbid water with poor visibility. Traditional frame-based imaging systems often struggle in these conditions due to motion blur, limited temporal resolution, and sensitivity to lighting fluctuations. Event cameras, with their high temporal resolution, low latency, high dynamic range and comparatively lower power consumption\cite{ebcsurveyscaramuzz}, offer a promising alternative for robust visual sensing in such scenarios.

Event-based vision is a bio-inspired paradigm where sensors asynchronously record changes in brightness at each pixel, rather than capturing frames at fixed intervals. This results in a stream of events that encode motion and scene dynamics with exceptional temporal fidelity. While event cameras have demonstrated success on terrestrial and aerial robotics platforms \cite{zhu2023eventcamerabasedvisualodometry, bhattacharya2024monoculareventbasedvisionobstacle, 9292606}, their application in underwater domains remains relatively underexplored. Recent efforts have begun to highlight their potential for underwater perception, particularly in tasks involving high-speed motion or variable lighting, such as marine life monitoring or navigating autonomous underwater vehicles (AUVs) in complex environments\cite{zhangeventbasedauvdocking}\cite{takatsukaplanktonmonitoringebc}. 
\begin{figure}[t]
    \centering
    \includegraphics[width=1\linewidth]{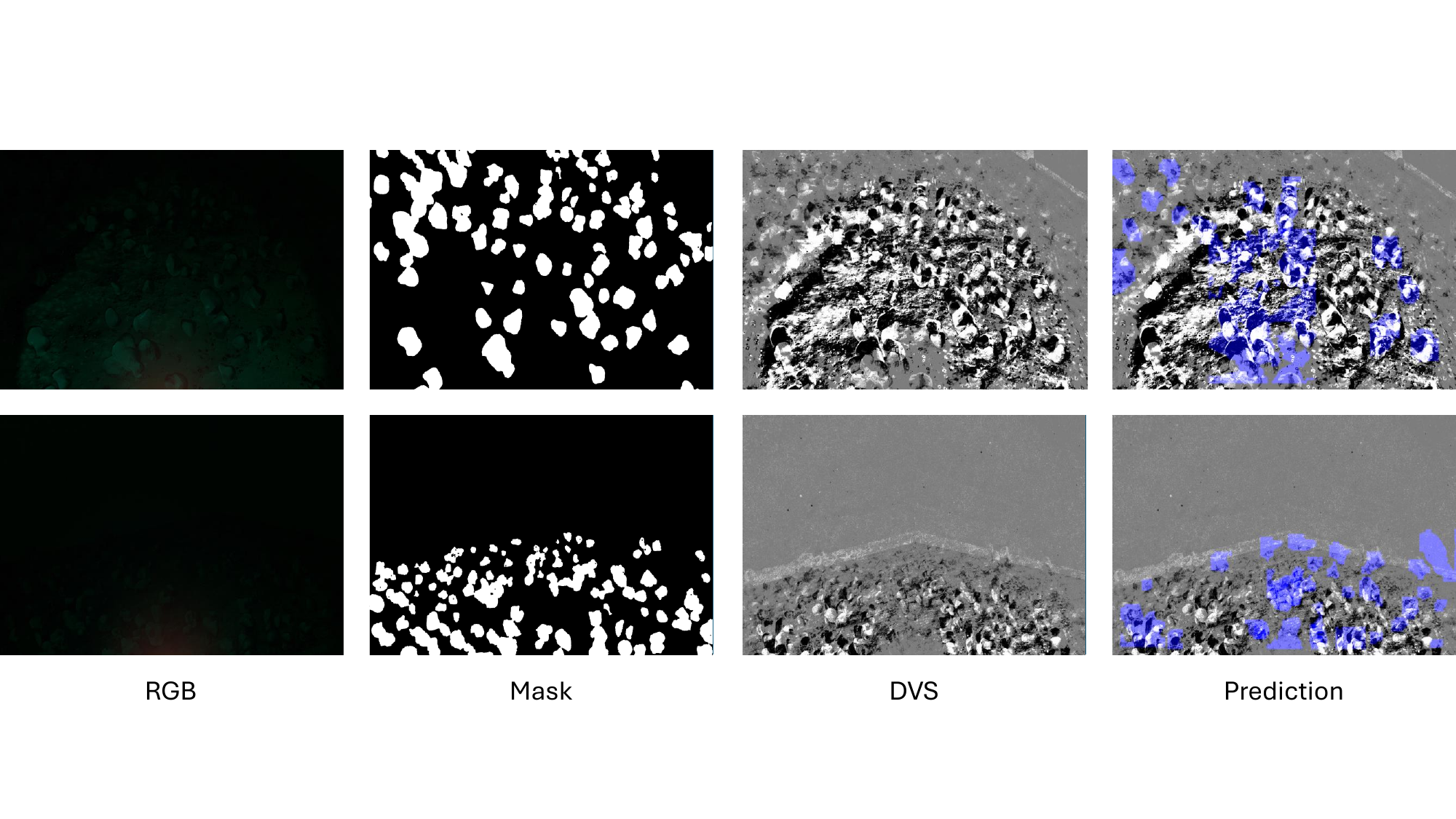}
    \caption{Illustration of EREBUS frame-by-frame image processing}
    \label{fig:EREBUS_output}
\end{figure}
While event cameras offer distinct advantages, we do not propose them as a wholesale replacement for conventional frame-based systems. Instead, we advocate for their complementary integration, especially in scenarios with rapid motion or low-light conditions, where event data can enhance perception capabilities and enable faster, more robust inference. While there is extensive literature and datasets available for vision sensors in the underwater domain, the event based paradigm remains under served in this regard. To address both this and the previously mentioned points, we make the following novel contributions.

\begin{itemize}
 \item \textbf{Simulation pipeline for underwater event vision:}
We present a fully functional pipeline that utilizes the physics based underwater simulation capabilities of Blender\cite{blender} along with event stream generation and downstream object detection. 

 \item \textbf{Demonstration of event cameras’ potential for underwater environments:}
We demonstrate, through simulation and detection performance, that event-based sensing is highly promising for challenging underwater conditions.

 \item \textbf{Few-shot segmentation from simulated event data:}
We show that realistic, physics-based simulations can be used to train a YOLO model with only a few annotated samples — highlighting the potential of synthetic event data for bootstrapping robust perception models with minimal labeling effort.

 \item \textbf{Extensible simulation framework for marine robotics:}
Our Blender-based simulation setup is easily modifiable and scene-agnostic — it can easily be adapted to simulate coral reefs, shipwrecks, marine life, or environmental monitoring scenarios. This enables broad applicability for developing and testing event-based models without requiring expensive real-world underwater data collection.

 \item \textbf{Foundational groundwork for a public dataset and benchmark:}
We aim to extend this pipeline into an open-source tool and potentially release a benchmark dataset of underwater event streams — supporting future research in event-based underwater robotics.
\end{itemize}

\begin{figure*}[ht]
    \centering
    \resizebox{1\textwidth}{!}{%
        \includegraphics{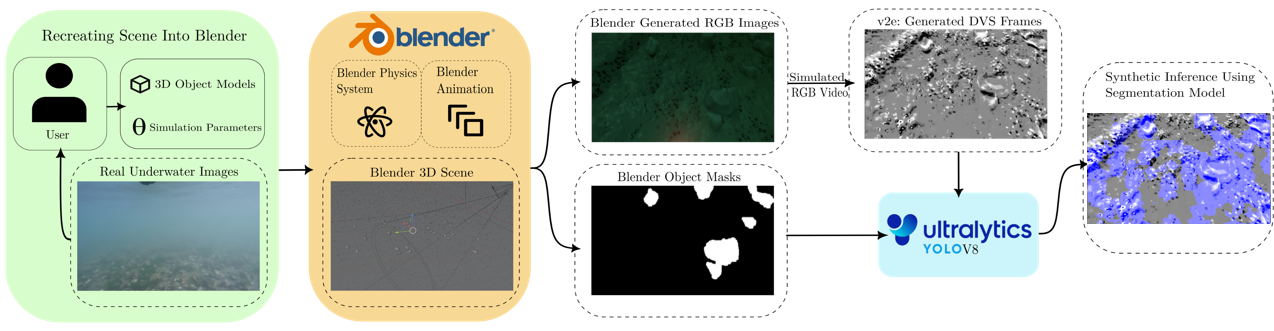}
    }
    \caption{A depiction of the pipeline of EREBUS. The Blender generated RGB image was brightened for demonstration purposes.}
    \label{fig:pipeline}
\end{figure*}

%% file: sections/related_work.tex
\subsection{Underwater Simulations}
Owing to the logistical difficulties encountered in obtaining data in the underwater domain, simulations have proven to be a reliable method to generate convincing training data for vision models and a testing ground for underwater robotics algorithms\cite{10406819amer, 9812353potokar,song2025oceansimgpuacceleratedunderwaterrobot}. Similar simulations have also been built around specific underwater tasks, such as oyster detection and counting on the seabed\cite{lin22oystersim,lin2025odysseeoysterdetectionyielded,wei2024unsupervised}. These simulations are engineered to address certain challenges inherent to the underwater domain, which are expanded on below.

\subsection{Challenges of the Underwater Domain}
The underwater domain is a common avenue for robotics applications, with AUVs being used for structural health monitoring, mine detection and oceanic exploration\cite{wibisonosurveyuuv}\cite{WALDNER2024100112}. However, deployed AUVs commonly face the issue of poor visibility and high occlusion. Several efforts have been undertaken to circumvent these common roadblocks. \cite{roserunderwtaervisibilitassessment} and \cite{zhou2022light} utilize light attenuation models to estimate the visibility properties of the occluded image, while \cite{weiyinotchessbutunderwater} leverages the light polarization information contained within the Stokes vector as the foundation to create a polarimetric imaging model. However, underwater lighting has been shown to adversely affect marine life by several studies\cite{stanton2024effects, geoffroy2021pelagic, kochevar1998effects}, thus to ensure the ecological soundness of AUV applications underwater, alternative strategies must be considered.

\subsection{Event Based Cameras}
While these works present viable avenues for mitigating the issues faced underwater, event based cameras also present a promising alternative. Unlike conventional frame-based cameras that record entire images at fixed intervals, event-based cameras operate asynchronously at the pixel level — recording events only when a change in brightness is detected. This affords them several advantages; namely their superior temporal resolution, high dynamic range and low power consumption\cite{ebcsurveyscaramuzz}. Extensive prior work has been done in utilizing their properties for robotics, such as for live obstacle detection and avoidance for quadrotors\cite{sanket2020evdodgenetdeepdynamicobstacle} and for live reconstruction of 3-D scenery during high speed egomotion\cite{xiong2024event3dgseventbased3dgaussian}. Additionally, their applications have been studied in self-driving cars \cite{8578666scaramuzz}, and to quadrupedal robots\cite{10161392}. Despite the volume of literature pertaining to their applications, event based cameras remain underutilized in the underwater domain. 


\subsection{Applications of Event Cameras to Underwater Domain}
Recent research has explored event-based vision for various underwater applications. For AUV docking, event cameras paired with spiking neural networks have improved target detection and docking precision in low-visibility settings\cite{zhangeventbasedauvdocking,maass1997networks}. Event data has also been fused with RGB imagery to enhance underwater images by reducing haze, uneven lighting, and color distortion\cite{bi2024rgb}. Additionally, event cameras have enabled high-resolution flow visualization by tracking laser-illuminated particles, aiding fluid dynamics analysis in environmental and medical studies\cite{willert2022event}.

In addition, collision avoidance systems\cite{dalmas2024review}\cite{Dadson_2025_WACV} for Autonomous Underwater Vehicles (UUVs) are beginning to incorporate event-based sensors. By integrating visual event data with auditory cues, these systems can perform real-time obstacle detection and improve the autonomy and safety of underwater operations in complex or obstacle-rich settings. 

This does not address the inherent difficulties in acquiring real underwater data; simulations can mitigate this issue by providing an avenue for the generation of convincing event based imagery.

\subsection{Event Based Simulations}
Simulation pipelines dedicated towards event-based imagery are varied in approach. Earlier works have primarily focused on the logarithmic intensity differences between successive frames \cite{8296630Bi} \cite{7862386kaiserjacq}, but this approach struggles with fast moving scenes. More recent techniques like EventGAN \cite{zhu2019eventganleveraginglargescale} leverage a generative adversarial network to generate convincing synthetic event data. Additionally, full 3D simulations of visual scenes are used, such as ESIM\cite{gehringrebecqesim}, an OpenGL based simulation that takes in a scene with a moving camera, rendering the changes in brightness as the camera moves along its trajectory. \cite{Mueggler_2017} leverages the DAVIS (Dynamic and Active-pixel Vision Sensor) in a wide domain of synthetic and real environments for visual odometry and SLAM. 

These simulators have laid the groundwork for event data generation but they are not specific to the underwater domain. In contrast, our work introduces a dedicated simulation and perception pipeline tailored for event-based vision in marine environments.

%% file: sections/methods.tex
In this work, we present a pipeline for simulating event-based data in underwater environments and using it for downstream perception tasks, such as object segmentation. Our methodology is designed to replicate realistic underwater scenes using Blender, generate synthetic event streams via an event camera simulator, and train lightweight neural models for effective object detection with limited supervision. The following subsections detail each stage of our pipeline.

\subsection{Underwater Scene Simulation in Blender}
We first construct a synthetic underwater environment using Blender. The scene consisted of a random distribution of rocks on the seabed, with thousands of suspended particles, to simulate a relatively occluded shallow-water domain. These particles introduce natural visual noise, mimicking conditions in which suspended solids or plankton may affect visibility and contrast in marine settings. To simulate realistic motion, a virtual camera is animated to traverse the scene, with the intent of emulating the movement of an autonomous underwater vehicle (AUV) navigating through the environment. The scene features limited visibility, with a spotlight mounted to the camera being the only source of illumination in the scene. The Blender engine is used to render a sequence of RGB images in 1920x1080 resolution, along with the corresponding segmentation masks of the rocks on the seafloor, capturing the temporal evolution of the scene as observed by the moving camera. 

\subsection{Event Stream Generation}
The rendered image sequence is then processed using an event camera simulator. This simulator models the behavior of a Dynamic Vision Sensor (DVS), converting intensity images into asynchronous event streams. Each event is triggered based on a change in pixel-level brightness, producing a time-ordered sequence of (x, y, polarity, timestamp) events. This stream is visualized as a DVS video, representing the kind of output a real event camera would generate under similar conditions\cite{hu2021v2e}.

\subsection{Object Segmentation using YOLO with Few-Shot Learning}
To demonstrate the utility of the simulated event data, we train a custom object detection and segmentation model using the YOLO (You Only Look Once) architecture\cite{redmon2016lookonceunifiedrealtime}. Specifically we utilize the YOLOv8-n, the smallest available model by weight count, as it is the most suited to be mounted on an edge device, where limited storage capacity is a concern. The DVS video frames are used as input to the model, and a small annotated dataset is sufficient to train YOLOv8 to reliably detect rock formations in the scene. This few-shot learning approach leverages the spatiotemporal richness of event data and highlights its efficiency in low-data regimes as shown in Fig.~\ref{fig:EREBUS_output}. We obtained a maximum mAP (Mean Average Precision) of 0.83 on a training set of 10 images over a similarly sized validation set for 250 epochs.  Despite the small size of the training set, we obtained impressive results for the live inference over the synthetic DVS footage, sporting a large number of detections with high confidence intervals. 

\subsection{Enhancing Robustness through Simulated Particle Noise and Adaptive Training}
Building upon the object detection capabilities demonstrated in the previous subsection, we further explore the robustness of our YOLO-based segmentation model under conditions that more closely resemble real-world underwater environments. A major challenge in such settings is particle noise caused by suspended sediments, bubbles, or other particulate matter, which can significantly distort visual inputs and reduce detection reliability.

\begin{figure}
    \centering    
    \includegraphics[width=1\linewidth]{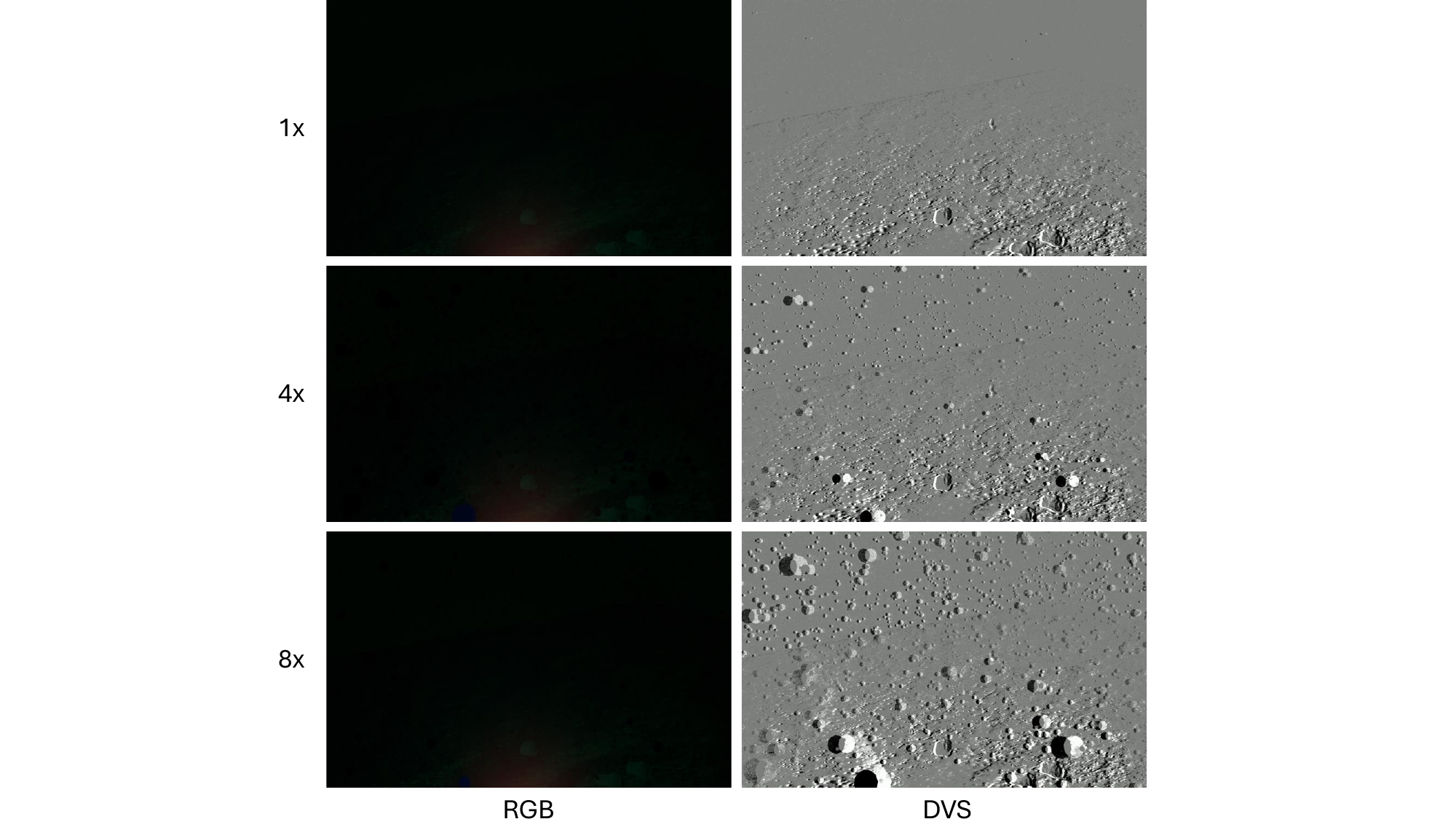}
    \caption{Illustration showing particle sizes ranging from the default to 8× larger. Best viewed at 400\% zoom. Note: the background is not completely black.}
    \label{fig:Particle_size}
\end{figure}

To emulate these conditions, we introduce varying levels of particle noise in our simulation by systematically increasing both the size and the number of particles. Our observations show that when the particle size exceeds twice the baseline configuration, the resulting DVS output becomes heavily affected. This degrades object detection performance due to the increase in background clutter and occlusion effects in the event stream as shown in Fig.~\ref{fig:Particle_size}

Crucially, because the simulation environment allows fine-grained control over such disturbances, we are able to incorporate particle noise into the training process itself. By exposing the model to noisy conditions during training, we enable it to learn noise-invariant features, improving its ability to detect target objects (e.g., rock formations) despite visual interference. This customized training strategy enhances the model's robustness and further demonstrates the advantage of combining synthetic DVS data with few-shot learning.

\subsection{Proposed Pipeline Summary}
We propose this simulation-to-perception pipeline as a foundational framework for developing and evaluating event-based vision algorithms in underwater robotics. While we focused on rock counting for this iteration of the simulation, the approach is highly flexible and could be capable of supporting a wide range of marine tasks. Our preliminary results indicate that even with minimal supervision, the pipeline enables effective segmentation from event-based inputs — validating its potential for real-world use.

%% file: sections/future_work.tex
In this work, we propose a simulation framework to generate synthetic event data for marine scenes. The simulation is intended to replicate the scenario in which an AUV navigates a realistic underwater scene. We present preliminary results from our simulations, showcasing event stream outputs in representative marine scenarios. These results demonstrate the viability of event-based vision for underwater robotics, even without extensive noise modeling. Our goal is to develop high-fidelity event stream simulations tailored for underwater environments, serving as a foundation for training and benchmarking event-based perception algorithms in an underwater domain. By contributing to the emerging intersection of event-based vision and marine robotics, we aim to enable more robust, efficient, and adaptable underwater perception systems. We also envision this work paving the way for real-world deployments, closing the gap between simulation and field-ready marine sensing technologies.
\newline

A promising future direction lies in fusing event-based and traditional frame-based data streams. Such hybrid systems could capitalize on the strengths of both modalities, thus leveraging the structural richness of RGB frames and the temporal precision of event streams to enhance detection reliability and computational efficiency under challenging underwater conditions.
\newline

Looking ahead, we aim to broaden the applicability of the sim, by simulating a more diverse assortment of environments which are realistically encountered by underwater vehicles. Future iterations of our pipeline can incorporate physics-based noise modeling and more varied sources of water occlusion to further improve realism. Other tasks such as species monitoring could also potentially be addressed with a modified version of our simulation. Our future work consists of a comprehensive synthetic event based dataset which can be open-sourced and made available to all, as a means to improve the performance of mounted vision models on autonomous underwater vehicles in a wide range of commonly encountered domains. Extensive testing of the trained vision models will follow, on event-based data in diverse underwater domains.